\documentclass[conference]{IEEEtran}
\IEEEoverridecommandlockouts
\usepackage{cite}
\usepackage{amsmath,amssymb,amsfonts}
\usepackage{algorithmic}
\usepackage{graphicx}
\usepackage{textcomp}
\usepackage{xcolor}
\usepackage{cite}
\usepackage{amsmath,amssymb,amsfonts}
\usepackage{algorithmic}
\usepackage{geometry}
\usepackage{textcomp}
\usepackage{booktabs}
\usepackage{subfigure}
\usepackage{framed}
\usepackage{wasysym}
\usepackage{array}
\usepackage[square,sort,comma,numbers]{natbib}
\usepackage{multirow}
\def\BibTeX{{\rm B\kern-.05em{\sc i\kern-.025em b}\kern-.08em
    T\kern-.1667em\lower.7ex\hbox{E}\kern-.125emX}}
\begin{document}

\title{Emoji-based Fine-grained Attention Network for Sentiment Analysis in the Microblog Comments*\\
{\footnotesize \textsuperscript{*}Note: Sub-titles are not captured in Xplore and
should not be used}
\thanks{Identify applicable funding agency here. If none, delete this.}
}

\author{\IEEEauthorblockN{1\textsuperscript{st} Deng Yang}
\IEEEauthorblockA{\textit{Xihua University} \\
Chengdu, China \\
212020081200025@stu.xhu.edu.cn}
\and
\IEEEauthorblockN{2\textsuperscript{nd} Liu Kejian}
\IEEEauthorblockA{\textit{Xihua University} \\
Chengdu, China \\
liukejian@gmail.com}
\and
\IEEEauthorblockN{3\textsuperscript{rd}Yang Cheng}
\IEEEauthorblockA{\textit{Xihua University} \\
Chengdu, China \\
212020085400097@stu.xhu.edu.cn}

}
\maketitle

\begin{abstract}
Microblogs have become a social platform for people to express their emotions in real-time, and it is a trend to analyze user emotional tendencies from the information of Microblogs. The dynamic features of emojis can affect the sentiment polarity of microblog texts. Since existing models seldom consider the diversity of emoji sentiment polarity,the paper propose a microblog sentiment classification model based on ALBERT-FAET. We obtain text embedding via ALBERT pretraining model and learn the inter-emoji embedding with an attention-based LSTM network. In addition, a fine-grained attention mechanism is proposed to capture the word-level interactions between plain text and emoji. Finally, we concatenate these features and feed them into a CNN classifier to predict the sentiment labels of the microblogs. To verify the effectiveness of the model and the fine-grained attention network, we conduct comparison experiments and ablation experiments. The comparison experiments show that the model outperforms previous methods in three evaluation indicators (accuracy, precision, and recall) and the model can significantly improve sentiment classification. The ablation experiments show that compared with ALBERT-AET, the proposed model ALBERT-FAET is better in the metrics, indicating that the fine-grained attention network can understand the diversified information of emoticons.
\end{abstract}


\section{Introduction}
With the rapid development of the Internet, microblog posts have become a platform for young users to express their opinions. Sentiment analysis is the process of analyzing, processing, generalizing and reasoning about subjective texts filled with emotional expression, which has attracted much attention in natural language processing. Microblog comment texts are more informal than ordinary texts, and to analyse the sentiment of microblog comments will generate much practical value. For example, it can be used for e-commerce platforms to conduct microblog marketing and make personal recommendations for users. It can also be used to monitor online public opinion, grasp people's opnions and emotions about social events. In addition, it can understand the public’s mental health and identify potential patients with depression nad anxiety.

Traditional methods mainly construct sentiment dictionaries to accomplish the sentiment classification task. Based on manually established seed adjective vocabularies, Hu and Liu \cite{hu2004mining} proposed a bootstrapping technique by using WordNet to predict the sentiment tendency of opnion words.

Deep learning has achieved good performance in many natural language processing tasks in the past few years. The sentiment classification task has succeeded as a subtask in natural language processing. Zhang \cite{zhang2016study} conducts recurrent neural networks to obtain word semantic features and word sequence features from sentence vectors and feed them into a softmax classifier to predict the sentiment label of each sentence in Chinese microblogs. Chen \cite{chen2020exploration} et al. combined military sentiment lexicon and BiLSTM model to boost the accuracy and F1-measure. Zhang \cite{yangsen2018microblog} also used the BiLSTM model to encode semantic information of text, combined with sentiment symbol library to enhance sentiment analysis.
The existence of the same emoji expressing different emotions in different scenarios in Weibo comments complicates the task of sentiment classification. For example, "My stomach hurts, I don't want to talk \includegraphics[scale=0.2]{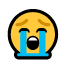}"   expresses a negative sentiment in the context. A different scenario, "The clothes I ordered arrived and they look beautiful \includegraphics[scale=0.2]{cry.png}," expresses the exact opposite sentiment compared with the native sentiment polarity in the context. At the same time, the number of emojis impacts the sentiment polarity of the sentence. For example, "I do not want to say anything \includegraphics[scale=0.2]{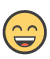} \includegraphics[scale=0.2]{smile.png} \includegraphics[scale=0.2]{smile.png}," multiple emojis  \includegraphics[scale=0.2]{smile.png} strengthen the negative emotion expression. Therefore, with the help of sentiment words that co-occur with emojis, we need to extract the necessary textual or contextual features to establish certain connections between emoji and plain text.

\begin{table*}[]
	\centering
	\caption{Microblog comments with emojis.}
	\begin{tabular}{ c|m{1.5cm}<{\centering}|l}
		\hline
		emoji & sentiment & microblog comments                                     \\ \hline
		\multirow{2}{*}{\includegraphics[scale=0.2]{cry.png}}& positive  & The clothes I ordered arrived and they look beautiful. \includegraphics[scale=0.2]{cry.png}       
		\\ \cline{2-3} 
		& negative  & My stomach hurts, I don't want to talk.\includegraphics[scale=0.2]{cry.png}                 \\ \hline
		\multirow{2}{*}{\includegraphics[scale=0.2]{smile.png}}& positive  & It is a nice day, I feel good.\includegraphics[scale=0.2]{smile.png} \\ \cline{2-3} 
		& negative  & I do not want to say anymore.\includegraphics[scale=0.2]{smile.png} \includegraphics[scale=0.2]{smile.png}\includegraphics[scale=0.2]{smile.png}             \\ \hline
	\end{tabular}
\end{table*}

Most of the existing methods conduct coarse-grained mechanisms to capture interactions between emoji and plain text. If emoji in a complex network environment present emotional polarity diversification or if there are multiple emoji in a sentence. Therefore, this paper proposes a fine-grained attention mechanism to capture the interaction between emoji and plain text. The main problem of the research is to analyze the sentiment tendency information of the microblog text. The main contributions of this paper are summarized as follows:

1. We use ALBERT pre-trained model to learn the word vector of microblog comments. Simultaneously, the model is easy to deploy in engineering due to its fewer parameters.

2. We first adopt emoji2vec to learn bi-sense emoji embeddings and then obtain the inter-emoji embedding as a weighted average of the bi-sense emoji embedding base on the attention mechanism.

3. We propose a fine-grained attention mechanism to extract word-level interaction information between emoji and plain text.

4. Morever,we design an emoji alignment loss in the objective function to boost the difference of the attention weights towards the emoji which have the same text and different sentiment polarity.
\section{Related Work}
\subsubsection{Sentiment Anaylysis}
Sentiment analysis is the process of analyzing, processing, generalizing and reasoning about subjective texts filled with emotional expression, which has attracted much attention in natural language processing. Mingjie Ling  \cite{ling2020hybrid} extracted word representation and sentence position representation from multichannel CNN and LSTM respectively, and experiments showed that the model achieved a better polarity classification ability for Chinese Weibo. Duyu Tang  \cite{tang2015effective} proposed a target-dependent LSTM model where target signals are taken into consideration to boost the classification accuracy.

Mingjie Ling  \cite{ling2020hybrid} extracted word representation and sentence position representation from multichannel CNN and LSTM respectively, and experiments showed that the model achieved a better polarity classification ability for Chinese Weibo. Duyu Tang  \cite{tang2015effective} proposed a target-dependent LSTM model where target signals are taken into consideration to boost the classification accuracy.

The earliest application of attention mechanisms is in the field of computer vision. In the literature \cite{mnih2014recurrent}, researchers adopt the attention mechanism on RNN models to implement image classification. Then, Bahdanau et al \cite{bahdanau2014neural}conducted the attention mechanism to machine translation tasks, which means that the attention mechanism has attracted a lot of attention in the natural language processing. In 2017, the Google machine translation team \cite{vaswani2017attention} built the whole model framework with the Attention mechanism to replace the traditional RNN method. It contains a fully connected feed-forward network between each layer in the encoder and decoder structure.Feifan Fan \cite{fan2018multi} proposed a fine-grained attention mechanism to capture word-level interactions between contexts and aspects in Twitter, and experiments demonstrate that the approach can effectively improve performance.

\subsubsection{Sentiment Analysis with Emoji}
Li Nan \cite{LiNan2021} explored the distribution characteristics and sentiment transformation regularities of emojis in microblog comments from multiple perspectives. The paper classified emojis into high sentiment stability and low sentiment stability based on thresholds and experiment confirmed that emojis can be used in opinion  analysis to achieve more accurate sentiment classification. Novak P \cite{kralj2015sentiment} constructed a sentiment lexicon containing 751 emojis, and experiments demonstrated that comments with emoji expressed more positive sentiments than those without emoji. Pohl \cite{2017Beyond} investigated the similarity problem of emoji in terms of emoji keywords and emoji embeddings.Experiments verified the model's effectiveness in capturing associations between each emoji. Ben Eisner \cite{eisner2016emoji2vec}trained emoji and corresponding descriptive textual information to propose an emoji2vec model for emoji pre-trained embeddings.

Some researchers adopted deep learning to study the sentiment classification task of the emoji-based microblog comments. Zhao \cite{ZhaoXiaofang2020} adopted CNN and RNN networks based on attention mechanism to extract semantic features and weighted the sentiment tendency values of the text and emojis to predict the sentiment tendency of Chinese microblog comments. Felbo \cite{felbo2017using} predicted the appearance of emoji with pre-training deep neural networks, which is effective to extract emotion information from emoji in sentiment classification and sarcasm detection task. Li \cite{li2017joint} conducted a convolutional neural network to predict the occurrence of emoji and learn emoji embedding jointly through a matching layer based on cosine similarity. These approach, in our context, adopt emoji as independent inputs to predict the sentiment label, which suffer from ignoring the interactivity between emoji and plain text.

Lou \cite{lou2020emoji} combined attention mechanisms to measure the contribution of each word in sentiment polarity based on emojis, although the approach cannot effectively handle microblog comments containing multiple types of emojis. Yuan X \cite{yuan2021emoji} proposed an emoji-based collaborative attention network to learn the interactive sentiment semantics of text and emoji. The model feed the text vector, text-based emoji vector, and emoji-based text vector into the convolutional neural network to predict the sentiment polarity of microblog texts. Experimental results show that the method outperforms several baselines for microblog sentiment classification. Chen \cite{chen2018twitter} combines a more robust and fine-grained bi-sense emoji embedding to represent complex semantic and sentiment information effectively. An attention-based mechanism of the LSTM network selectively attend on the relevant emoji embeddings to understand rich semantics and sentiment better. The experiments on the Twitter dataset demonstrate the model outperforms the state-of-the-art models.

\section{THE PROPOSED MODEL}
\begin{figure*}[htbp]
	\centering
	\includegraphics[width=\linewidth,scale=0.3]{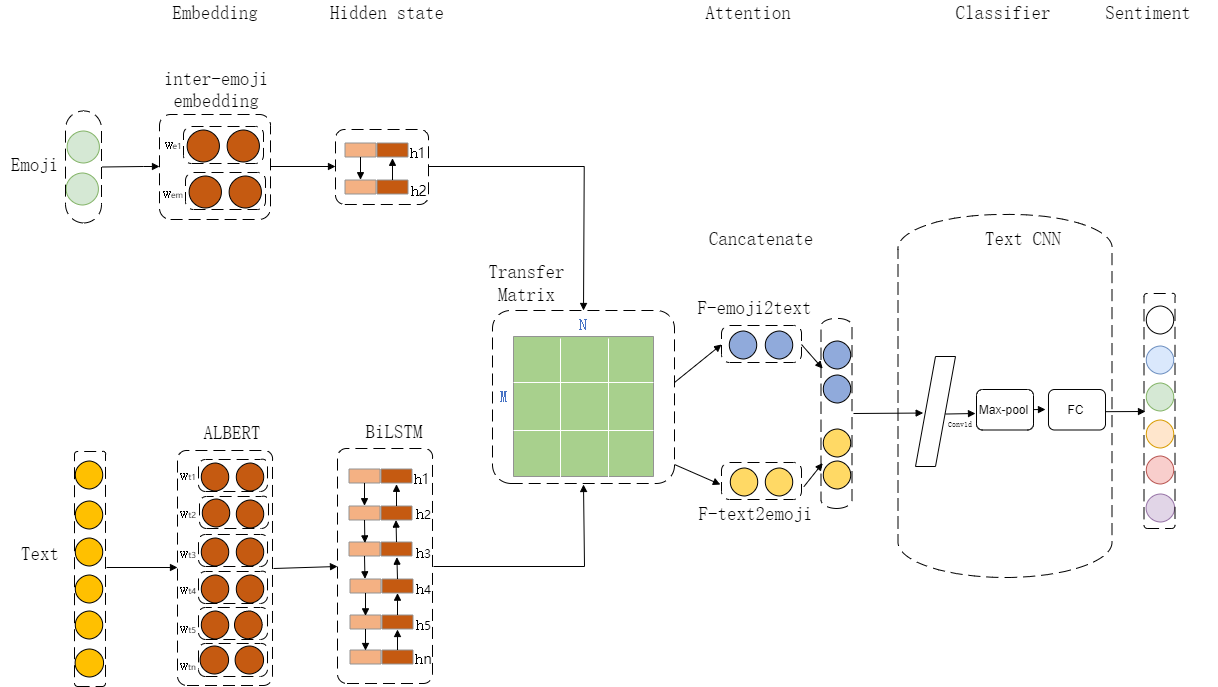}
	\caption{ALBERT-FAET Model}
\end{figure*}
The main problem to be solved in this study is the analysis
of the emotional information of barrage comments. For an given sentence $L^{a}=\left\{\boldsymbol{x}_{1} \boldsymbol{x}_{2}, \boldsymbol{x}_{3}, \ldots, \boldsymbol{x}_{n},\right. \\
\left. \boldsymbol{e}_{1},\boldsymbol{e}_{2}, \boldsymbol{e}_{3}, \ldots, \boldsymbol{e}_{n}\right\}            $
,where n represents the number of text words and m represents the number of emoji words.The sentiment
polarity $y_{i} \in\{1,0\}$ corresponds to 'positive'
and 'negative' sentiments of a reviewer.

We present the overall architecture of the proposed Emoji-based Fine-grained Attention Network model in Figure 4. It consists of the embedding layer, hidden layer, attention layer, and textCNN classifier layer.

\subsection{Embedding Layer}
\subsubsection{Text}
The text adopts ALBERT to learn word embedding. ALBERT utilizes factorized embedding parameterization, cross-layer parameter sharing, and sentence order prediction(SOP) strategies to deepen the model while reducing parameters, to achieve better results than the BERT models in various natural language processing tasks.
\subsubsection{Emoji}
We first assigned two distinct tokens to each emoji, one is the specific emoji used in a positive sentimental context, and the other is the emoji used in a negative sentimental context. Each token is embedded into a different vector using  emoji2vec to obtain a bi-sense embedding \cite{chen2018twitter} to each emoji. We first learn text embedding using ALBERT and obtain inter-emoji embedding by a simple attention network between plain text and emojis.
We feed the text and emoji into the embedding layer, the output sequence is $L^{b}=\left\{\boldsymbol{x}_{1} \boldsymbol{x}_{2}, \boldsymbol{x}_{3}, \ldots,\boldsymbol{x}_{n}, \boldsymbol{e}_{1}, \boldsymbol{e}_{2}, \ldots, \boldsymbol{e}_{m}\right\}$

\begin{figure}[htbp]
	\centering
	\includegraphics[scale=0.5]{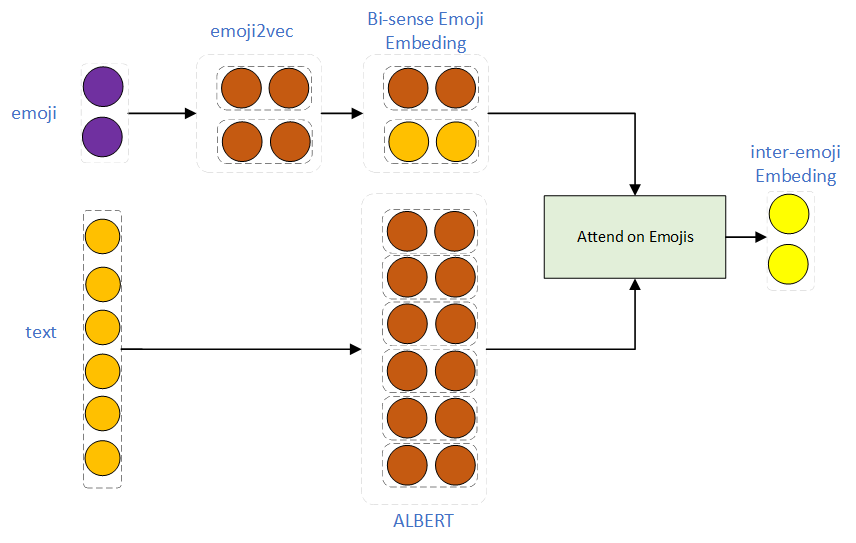}
	\caption{Emoji processing figure}
\end{figure}
\begin{equation}
	\begin{aligned}
		u_{t, i} &=f_{a t t}\left(\mathbf{e}_{t, i}, \mathbf{w}_{t}\right) 
	\end{aligned}
\end{equation}
\begin{equation}
	\begin{aligned}
		\alpha_{t, i} &=\frac{\exp \left(u_{t, i}\right)}{\sum_{i=1}^{m} \exp \left(u_{t, i}\right)} 
	\end{aligned}
\end{equation}

\begin{equation}
	\begin{aligned}
		\mathbf{v}_{t} &=\sum_{i=1}^{m}\left(\alpha_{t, i} \cdot \mathbf{e}_{t, i}\right)
	\end{aligned}
\end{equation}
$i \in(1, m)$ in $e_{t, i}$  denotes the i-th sense embedding of the emoji $e_{t}(m=2)$,$f_{a t t}\left(\cdot, \mathbf{w}_{t}\right)$ denotes the attention function that is based on the current word embedding, $\alpha_{t}$ represents the attention weight, and $\alpha_{t}$ represents the attention weight, and $\mathbf{v}_{t}$ represents the inter-emoji embedding.
\subsection{hidden Layer}
We adopts bidirectional Long Short-Term Memory Network(BiLSTM) to capture the temporal interactions among words. The operations in an LSTM unit for time step t is formulated in Equation below:
\begin{equation}
	\begin{aligned}
		\mathbf{i}_{t} &=\sigma\left(W_{i} \mathbf{x}_{t}+U_{i} \mathbf{h}_{t-1}+b_{i}\right) 
	\end{aligned}
\end{equation}
\begin{equation}
	\begin{aligned}
		\mathbf{f}_{t} &=\sigma\left(W_{f} \mathbf{x}_{t}+U_{f} \mathbf{h}_{t-1}+b_{f}\right) 
	\end{aligned}
\end{equation}
\begin{equation}
	\begin{aligned}
		\mathbf{o}_{t} &=\sigma\left(W_{o} \mathbf{x}_{t}+U_{o} \mathbf{h}_{t-1}+b_{o}\right) 
	\end{aligned}
\end{equation}
\begin{equation}
	\begin{aligned}
		\mathbf{g}_{t} &=\tanh \left(W_{c} \mathbf{x}_{t}+U_{c} \mathbf{h}_{t-1}+b_{c}\right) 
	\end{aligned}
\end{equation}
\begin{equation}
	\begin{aligned}
		\mathbf{c}_{t} &=\mathbf{f}_{t} \odot \mathbf{c}_{t-1}+\mathbf{i}_{t} \odot \mathbf{g}_{t} 
	\end{aligned}
\end{equation}
\begin{equation}
	\begin{aligned}
		\mathbf{h}_{t} &=\mathbf{o}_{t} \odot \tanh \left(\mathbf{c}_{t}\right)
	\end{aligned}
\end{equation}
where $\mathbf{h} t$ and $\mathbf{h}_{t-1}$ represent the current and previous hidden states, $\mathbf{x}_{t}$ denotes the current LSTM input, W and U denote the weight matrices.
Then,we extract deep semantic meaning from the sequence got from the embedding layer to obtain the semantic-rich feature vector $\dot{L^{c}}=\left\{\boldsymbol{T}_{1}, \boldsymbol{T}_{2}, \ldots, \boldsymbol{T}_{n}, \boldsymbol{E}_{1}, \boldsymbol{E}_{2}, \ldots, \boldsymbol{E}_{m}\right\}$, where $T_{i}$ is the feature vector of the first i text word after processing, and $E_{i}$ is the feature vector of the corresponding emoji.
\subsection{attention Layer}
In this paper, we propose a fine-grained attention mechanism \cite{munikar2019fine} to describe word-level interactions between emojis and text and to evaluate how emojis affect the overall sentence sentiment polarity.

Formally, we define an interaction matrix $U \in \mathbb{R}^{N * M}$ to describe the interaction between an emoji E and a text T, where $\mathrm{U}_{\mathrm{ij}}$ denotes the interactivity between the ith text word and the jth emoji. The interaction matrix U is computed by the following equation :

\begin{equation}
	U_{i j}=W_{u}\left(\left[E ; T_{j} ; E_{i} * T_{j}\right]\right)
\end{equation}

$W_{u} \in \mathbb{R}^{1 * 6 d}$  denotes the weight matrix, $[;]$ denotes the vector concatenation across row,* denotes elementwise multiplication, and then we use U to compute the attention vectors in both directions.

(1)F-Emoji2Text estimates which emoji should pay more attention to and are hence critical for determining the sentiment. We can compute the attention weights $\mathrm{e}^{\mathrm{ie}}$ by 
\begin{equation}
	\begin{aligned}
		s_{i}^{f e}=\max \left(U_{i,:}\right) 
	\end{aligned}
\end{equation}	
\begin{equation}
	\begin{aligned}
		e_{i}^{i e}=\frac{\exp \left(s_{i}^{f e}\right)}{\sum_{k=1}^{N} \exp \left(s_{k}^{f e}\right)}
	\end{aligned}
\end{equation}
where $S_{i}^{e}$ obtains the maximum similarity across column. And then we can get the attended vector as follows:
\begin{equation}
	m^{f e}=\sum_{i=1}^{N} e_{i}^{f e} \cdot E_{i}
\end{equation}
(2)F-Text2Emoji estimates which text should pay more attention to and are also critical for determining the sentiment. We can compute the attention weights $\mathrm{t}^{\mathrm{ft}}$ by 
\begin{equation}
	\begin{aligned}
		s_{i}^{f t}=\max \left(U_{i,:}\right)
	\end{aligned}	
\end{equation}

\begin{equation}
	\begin{aligned}		 
		t_{i}^{f t}=\frac{\exp \left(s_{i}^{f t}\right)}{\sum_{k=1}^{N} \exp \left(s_{k}^{f t}\right)}
	\end{aligned}
\end{equation}
Then we use an average pooling layer on to get the attended vector 
$m^{f t} \in \mathbb{R}^{2 d}$:
\begin{equation}
	m^{f t}=P_{\text {ooling }} g\left(\left[\boldsymbol{t}_{1}^{f t}, \ldots, \boldsymbol{t}_{i}^{f t}\right]\right)
\end{equation}
Finally,we concatenate fine-grained attention vectors as the final representation $m^{f t} \in \mathbb{R}^{4 d}$:
$\boldsymbol{m}=\left[\boldsymbol{m}^{f t} ; \boldsymbol{m}^{f e}\right]$
\subsection{textCNN Layer}
We take the concatenated vector into CNN classifier to predict the sentiment label of the microblog comments.

We adopt $\left[w_{1}, w_{2}, \ldots, w_{c}\right]$ to denote the set of filter kernels in the convolution operation and then map the input
$V \in \mathbb{R}^{d \times c}$ to a new feature map  $U \in \mathbb{R}^{d^{\prime} \times c^{\prime}}$.
\subsection{Model Training}
The existing methods train each text word and emoji separately, and seldom consider the fine-grained interaction between emoji and text. Experiments show that the fine-grained interaction between emoji and text can bring additional valuable information. Therefore, a text alignment loss function is proposed in the paper. The text is constrained by the alignment loss, and each text word will focus on the more important emoji by comparing with other text words.
\begin{equation}
	d_{i o}=\sigma\left(\boldsymbol{W}_{d}\left[\boldsymbol{T}_{i} ; \boldsymbol{T}_{o}\right]\right)
\end{equation}
\begin{equation}
	\ell_{\text { align }}=-\sum_{i=1}^{M-1} \sum_{o=i+1}^{M} \sum_{k=1}^{N} d_{i j} \bullet\left(x_{i k}^{f e}-x_{o k}^{f e}\right)^{2}
\end{equation}
In particular, for text words $\boldsymbol{x}_{i}$ and text words $\boldsymbol{x}_{o}$. The paper calculate the square loss on the fine-grained attention vectors  $\boldsymbol{x}_{i}^{f e}$ and $\boldsymbol{x}_{o}^{f e}$, and also estimate the distance$d_{i o}$ between $\boldsymbol{x}_{i}$ and $\boldsymbol{x}_{o}$ as the loss weight. Where $\sigma$ is the sigmoid function, $\boldsymbol{W}_{d}$ is the weight matrix for computing the distance, $\boldsymbol{x}_{i k}^{f e}$ and  $\boldsymbol{x}_{i k}^{f e}$ are the attention weights  on k-th context word towards text word $\boldsymbol{x}_{i}$ and $\boldsymbol{x}_{\circ}$ respectively.|

\section{Experiments}
In the paper, we crawled 60,000  microblog comments of length greater than 5 using API as the original experimental data. After removing the text excluding emoji from the microblogs and cleaning the data, the corpus are labelled for sentiment polarity by using a manual approach. Finally, we get a dataset consisting of 8930 texts containing emojis. Among them, 4418 were positive texts and 4512 were negative texts. In the paper, the texts are divided into training, validation and test based on 7:2:1.In detail,the training set includes 6250 sentences, the test set includes 1786 sentences and the validation set includes 894 sentences. The distribution of text  containing emojis is shown in the following table:

\begin{table}[h]
	\caption{The statistics of the datasets}
	\centering
	\begin{tabular}{cccc}
		\toprule
		Corpus & Positive & Negative & Total  \\ \midrule
		train  & 3092  & 3158       & 6250               \\
		test  & 884  & 902       & 1786          \\
		val  & 442  & 452       & 894          \\
		total  & 4418  & 4512      & 8930       \\	\bottomrule		
	\end{tabular}
\end{table}
\subsection{Metrics}
In the experiment, the evaluation indexes that have been commonly used in NLP tasks were adopted, and they were as follows: Precision(P), Recall (R), Accuracy (Acc), and F1 values, and they were respectively calculated by:

\begin{equation}
	\begin{gathered}
		\text { precision }=\frac{T P}{T P+F P} 
	\end{gathered}
\end{equation}
\begin{equation}
	\begin{gathered}
		\text { Recall }=\frac{T P}{T P+F N}
	\end{gathered}
\end{equation}

\begin{equation}
	\begin{gathered}
		\text { Accuracy }=\frac{T P+T N}{T P+T N+F N+F P}
	\end{gathered}
\end{equation}
Precision (P) indicates the proportion of samples correctly predicted. In other words, precision measures quality. Recall (R) represents the proportion of samples wrongly predicted. The f1-score is a number between 0 and 1, contributing to the measurement of precision and recall by calculating the harmonic mean of them. Accuracy (Acc)  represents the ratio between correctly predicted samples and the total number of samples, and it is a more global index.

\subsection{Hyperparameters}
In our Experiments, the hidden state d is set to 200, and the dropout ratio is set to 0.2 during the training period. The batch size is 64, the number of iterations is 10, and the maximum length of the text is limited to 100. The Adam optimizer was used to optimize model parameters, and the learning rate is initialized to 5*10-4. We randomly split the microblogs into the training, validation and test sets in the proportion of 7:2:1. The whole framework was bulit and trained by PyTorch.

\subsection{MODEL performance on Microblog Texts}
For this experiment, we test several state-of-the-art sentiment models on our dataset of microblog text:
$\mathbf{emoji2vec}$ contains 1661 embeddings, trained on Unicode descriptions of emojis, to improve natural language processing tasks that previously used word2vec to learn word embeddings.

\noindent $\mathbf{LSTM}$ \cite{hochreiter1997long} (long short-term memory ) is widely adopted in natural language processing tasks. It controls the transmission state using gate units and selectively memorizes information to process sequence tasks.

\noindent $\mathbf{TextCNN}$ \cite{chen2015convolutional}+word learns word embeddings through word2vec and feeds them into CNN networks to extract semantic features.

\noindent $\mathbf{TextRCNN}$ \cite{lai2015recurrent}+word replaces the convolutional layer with a bidirectional recurrent layer, and the concantenated vector is fed into the classifier to complete the classification compared with TextCNN.

\noindent $\mathbf{ET-BiLSTM}$ \cite{zhang2019deep} is an emojis-enhanced sentiment analysis model. The model contains the contextual information of the sentence into emoji to learn emoji-based auxiliary representation of the comments.

\noindent $\mathbf{BERT}$ \cite{devlin2018bert} is a pre-trained model proposed by Google in 2017, which adopts a masked language model to the bidirectional transformer to finish pre-training tasks. Then the last few layers of model parameters need to be fine-tuned to achieve satisfactory results.

\noindent $\mathbf{BERT+emoji2vec}$ uses BERT to learn the word vector of text and employs emoji2vec to learn emoji embedding. The model concatenates emoji embeddings and text embeddings to complete the sentiment classification.

\begin{table}[h]
	\caption{The performance comparisions of different models on the microblogs}
	\centering
	\begin{tabular}{@{}llllll@{}}
		\toprule
		Model & Acc & Micro-P & Micro-R &   \\ \midrule
		emoji2vec  & 0.658  & 0.642       & 0.660            &  \\
		TextCNN+word  & 0.745  & 0.741       & 0.743         &  \\
		ET-BiLSTM  & 0.821  & 0.819      & 0.823               &  \\
		BERT  & 0.802  & 0.814       & 0.806                &  \\
		BERT+emoji2vec  & 0.832  & 0.837       & 0.831         &  \\		
		ALBERT-FAET  & $\mathbf{0.852}$  & $\mathbf{0.855}$       & $\mathbf{0.856}$              &  \\ \bottomrule
		
	\end{tabular}
\end{table}

Compared with emoji2vec, TextCNN+word and BiLSTM+word models achieve good results in the sentiment classification task due to the neural networks' robust feature extraction ability. The results of the BERT model are generally better than those pre-trained with word2vec, indicating that BERT can capture deeper text features. Comparing BERT and BERT+emoji2vec, experiments show that emoji information can improve indexes effectively, indicating that emoji information can improve model performance. In addition, the ALBERT-FAET model proposed in this paper outperforms the previous benchmark on Chinese microblog comments. On the one hand,The model  assigns two distinct tokens to the emoji to obtain the bi-sense emoji embedding, and a text-based self-attention mechanism is adopted  to learn inter-emoji embeddings. On the other hand, the model proposes a fine-grained attention mechanism to capture the word-level interaction between emoji and text, which brings additional effective information.

\subsection{Analysis of ALBERT-FAET model}

\begin{table}[h]
	\caption{The performance comparisions of ALBERT-FAET variants}
	\centering
	\begin{tabular}{@{}llllll@{}}
		\toprule
		Model & Acc & Micro-P & Micro-R & Micro-R &  \\ \midrule
		ALBERT-AET  & 0.842  & 0.840       & 0.845       &  \\
		ALBERT-FAET  & $\mathbf{0.852}$  & $\mathbf{0.855}$     &$\mathbf{ 0.856 }$              &  \\ \bottomrule
	\end{tabular}
\end{table}

\begin{table*}[h]
	\caption{Prediction results of ALBERT-FAET and ALBERT-AET partial examples}
	\centering
	\makebox[\textwidth][c]{
		\begin{tabular}{@{}llllll@{}}
			\toprule
			Number & Microblog Text & ALBERT-FAET & ALBERT-AET & True Label &  \\ \midrule
			(1)  &My stomach hurts and I don't want to talk{\includegraphics[scale=0.2]{cry.png}}  & negative      & negative       &negative &  \\ \bottomrule
			(2)  &It's a good day and I feel happy{\includegraphics[scale=0.2]{smile.png}}  & positive      & positive       &positive & \\ \bottomrule
			(3)  &My favorite weather {\includegraphics[scale=0.2]{smile.png}}
			{\includegraphics[scale=0.28]{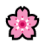}}
			{\includegraphics[scale=0.28]{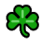}}  & positive      & positive       &positive  \\ \bottomrule
			(4)  &My ordered clothes arrived and they look beautiful {\includegraphics[scale=0.2]{cry.png}} & $\mathbf{positive}$      & negative       &positive  \\ \bottomrule
			(5)  &The favorite East King {\includegraphics[scale=0.2]{smile.png}} hung himself & $\mathbf{negative}$  & positive      &negative               &  \\ \bottomrule
			(6)&Yeah,What you say is relatively right{\includegraphics[scale=0.2]{smile.png}}
			{\includegraphics[scale=0.2]{smile.png}}
			{\includegraphics[scale=0.2]{smile.png}}  & $\mathbf{negative}$  & positive      & negative               &  \\ \bottomrule
			
		\end{tabular}
	}
\end{table*}
ALBERT-AET adopt a simple coarse-grained interaction between emoji and text to learn emoji features, and then  concatenates text features to finish the sentiment classification. ALBERT-FAET defines an interaction matrix to describe the word-level interaction between emoji and text. Emoticons in  (1), (2), and (3) have the same sentiment polarity as text in the current context, so ALBERT-AET, a coarse-grained attention mechanism, can correctly predict the sentiment polarity of the microblog text. The emoticons in (4) and (5) show inconsistency between emotion polarity and text polarity, and multiple emoticons appear in (6). At this time, the ALBERT-AET prediction results are very different from the real results. And ALBERT-FAET can accurately identify the sentiment polarity of the microblog text, which indicates that the fine-grained attention network can learn the dynamic feature information in the emoji, which is helpful to improve the sentiment classification accuracy of the model.

\subsection{Conclusion}
In this paper, we propose an emoji-based fine-grained attention network for microblog sentiment analysis. Specially, we propose an cross matrix to analyze the word-level interactions between text and emojis, which may bring extra valuable information. More importantly,we design a text alignment loss in the objective function to enhance the difference of the attention weights towards the text which have the same emoji and different sentiment polarities. The model achieves top results for sentiment classification on the crawled microblog dataset.

More fine-grained information can also be considered for sentiment analysis of the microblog comments. For instance, the sender's name of the microblog text, the time when the microblog text was sent mat taken into consideration.Besides,we may alter the finer granularity of the microblog text. We can label the microblog comments into five catorgories: very positive, positive, neutral, negative, and very negative.

In the future, we consider combining the multi-dimensional features of microblogs for sentiment analysis, such as considering the personality characteristics of microblog users.

\bibliographystyle{ieeetran}
\bibliography{myref}

\begin{thebibliography}{10}
\providecommand{\url}[1]{#1}
\csname url@samestyle\endcsname
\providecommand{\newblock}{\relax}
\providecommand{\bibinfo}[2]{#2}
\providecommand{\BIBentrySTDinterwordspacing}{\spaceskip=0pt\relax}
\providecommand{\BIBentryALTinterwordstretchfactor}{4}
\providecommand{\BIBentryALTinterwordspacing}{\spaceskip=\fontdimen2\font plus
\BIBentryALTinterwordstretchfactor\fontdimen3\font minus
  \fontdimen4\font\relax}
\providecommand{\BIBforeignlanguage}[2]{{%
\expandafter\ifx\csname l@#1\endcsname\relax
\typeout{** WARNING: IEEEtran.bst: No hyphenation pattern has been}%
\typeout{** loaded for the language `#1'. Using the pattern for}%
\typeout{** the default language instead.}%
\else
\language=\csname l@#1\endcsname
\fi
#2}}
\providecommand{\BIBdecl}{\relax}
\BIBdecl

\bibitem{hu2004mining}
M.~Hu and B.~Liu, ``Mining and summarizing customer reviews,'' in
  \emph{Proceedings of the tenth ACM SIGKDD international conference on
  Knowledge discovery and data mining}, 2004, pp. 168--177.

\bibitem{zhang2016study}
Y.~Zhang, Y.~Jiang, and Y.~Tong, ``Study of sentiment classification for
  chinese microblog based on recurrent neural network,'' \emph{Chinese Journal
  of Electronics}, vol.~25, no.~4, pp. 601--607, 2016.

\bibitem{chen2020exploration}
L.-C. Chen, C.-M. Lee, and M.-Y. Chen, ``Exploration of social media for
  sentiment analysis using deep learning,'' \emph{Soft Computing}, vol.~24,
  no.~11, pp. 8187--8197, 2020.

\bibitem{yangsen2018microblog}
Z.~Yangsen, Z.~Jia, H.~Gaijuan, and J.~Yuru, ``Microblog sentiment analysis
  method based on a double attention model,'' \emph{Journal of Tsinghua
  University (Science and Technology)}, vol.~58, no.~2, pp. 122--130, 2018.

\bibitem{ling2020hybrid}
M.~Ling, Q.~Chen, Q.~Sun, and Y.~Jia, ``Hybrid neural network for sina weibo
  sentiment analysis,'' \emph{IEEE Transactions on Computational Social
  Systems}, vol.~7, no.~4, pp. 983--990, 2020.

\bibitem{tang2015effective}
D.~Tang, B.~Qin, X.~Feng, and T.~Liu, ``Effective lstms for target-dependent
  sentiment classification,'' \emph{arXiv preprint arXiv:1512.01100}, 2015.

\bibitem{mnih2014recurrent}
V.~Mnih, N.~Heess, A.~Graves \emph{et~al.}, ``Recurrent models of visual
  attention,'' \emph{Advances in neural information processing systems},
  vol.~27, 2014.

\bibitem{bahdanau2014neural}
D.~Bahdanau, K.~Cho, and Y.~Bengio, ``Neural machine translation by jointly
  learning to align and translate,'' \emph{arXiv preprint arXiv:1409.0473},
  2014.

\bibitem{vaswani2017attention}
A.~Vaswani, N.~Shazeer, N.~Parmar, J.~Uszkoreit, L.~Jones, A.~N. Gomez,
  {\L}.~Kaiser, and I.~Polosukhin, ``Attention is all you need,''
  \emph{Advances in neural information processing systems}, vol.~30, 2017.

\bibitem{fan2018multi}
F.~Fan, Y.~Feng, and D.~Zhao, ``Multi-grained attention network for
  aspect-level sentiment classification,'' in \emph{Proceedings of the 2018
  conference on empirical methods in natural language processing}, 2018, pp.
  3433--3442.

\bibitem{LiNan2021}
L.~Nan and Z.~Yuhui, ``An analysis of web opinion combining the dynamic
  characteristics of emoji,'' \emph{Modern Intelligence}, 2021.

\bibitem{kralj2015sentiment}
P.~Kralj~Novak, J.~Smailovi{\'c}, B.~Sluban, and I.~Mozeti{\v{c}}, ``Sentiment
  of emojis,'' \emph{PloS one}, vol.~10, no.~12, p. e0144296, 2015.

\bibitem{2017Beyond}
H.~Pohl, C.~Domin, and M.~Rohs, ``Beyond just text: Semantic emoji similarity
  modeling to support expressive communication,'' \emph{ACM Transactions on
  Computer-Human Interaction}, vol.~24, no.~1, pp. 1--42, 2017.

\bibitem{eisner2016emoji2vec}
B.~Eisner, T.~Rockt{\"a}schel, I.~Augenstein, M.~Bo{\v{s}}njak, and S.~Riedel,
  ``emoji2vec: Learning emoji representations from their description,''
  \emph{arXiv preprint arXiv:1609.08359}, 2016.

\bibitem{ZhaoXiaofang2020}
Z.~Xiaofang and J.~Zhigang, ``Multi-dimensional sentiment classification of
  microblog based on emoticons and short texts,'' \emph{Journal of Harbin
  Institute of Technology}, vol.~52, no.~5, pp. 113--120, 2020.

\bibitem{felbo2017using}
B.~Felbo, A.~Mislove, A.~S{\o}gaard, I.~Rahwan, and S.~Lehmann, ``Using
  millions of emoji occurrences to learn any-domain representations for
  detecting sentiment, emotion and sarcasm,'' \emph{arXiv preprint
  arXiv:1708.00524}, 2017.

\bibitem{li2017joint}
X.~Li, R.~Yan, and M.~Zhang, ``Joint emoji classification and embedding
  learning,'' in \emph{Asia-Pacific Web (APWeb) and Web-Age Information
  Management (WAIM) Joint Conference on Web and Big Data}.\hskip 1em plus 0.5em
  minus 0.4em\relax Springer, 2017, pp. 48--63.

\bibitem{lou2020emoji}
Y.~Lou, Y.~Zhang, F.~Li, T.~Qian, and D.~Ji, ``Emoji-based sentiment analysis
  using attention networks,'' \emph{ACM Transactions on asian and low-resource
  language information processing (TALLIP)}, vol.~19, no.~5, pp. 1--13, 2020.

\bibitem{yuan2021emoji}
X.~Yuan, J.~Hu, X.~Zhang, H.~Lv, and H.~Liu, ``Emoji-based co-attention network
  for microblog sentiment analysis,'' in \emph{International Conference on
  Neural Information Processing}.\hskip 1em plus 0.5em minus 0.4em\relax
  Springer, 2021, pp. 3--11.

\bibitem{chen2018twitter}
Y.~Chen, J.~Yuan, Q.~You, and J.~Luo, ``Twitter sentiment analysis via bi-sense
  emoji embedding and attention-based lstm,'' in \emph{Proceedings of the 26th
  ACM international conference on Multimedia}, 2018, pp. 117--125.

\bibitem{munikar2019fine}
M.~Munikar, S.~Shakya, and A.~Shrestha, ``Fine-grained sentiment classification
  using bert,'' in \emph{2019 Artificial Intelligence for Transforming Business
  and Society (AITB)}, vol.~1.\hskip 1em plus 0.5em minus 0.4em\relax IEEE,
  2019, pp. 1--5.

\bibitem{hochreiter1997long}
S.~Hochreiter and J.~Schmidhuber, ``Long short-term memory,'' \emph{Neural
  computation}, vol.~9, no.~8, pp. 1735--1780, 1997.

\bibitem{chen2015convolutional}
Y.~Chen, ``Convolutional neural network for sentence classification,'' Master's
  thesis, University of Waterloo, 2015.

\bibitem{lai2015recurrent}
S.~Lai, L.~Xu, K.~Liu, and J.~Zhao, ``Recurrent convolutional neural networks
  for text classification,'' in \emph{Twenty-ninth AAAI conference on
  artificial intelligence}, 2015.

\bibitem{zhang2019deep}
J.~Zhang, X.~Li, Y.~Du, X.~Chen, Y.~Li, J.~Zheng, and Z.~Wang, ``A deep
  learning model enhanced with emojis for sina-microblog sentiment analysis,''
  in \emph{2019 IEEE International Conferences on Ubiquitous Computing \&
  Communications (IUCC) and Data Science and Computational Intelligence (DSCI)
  and Smart Computing, Networking and Services (SmartCNS)}.\hskip 1em plus
  0.5em minus 0.4em\relax IEEE, 2019, pp. 236--242.

\bibitem{devlin2018bert}
J.~Devlin, M.-W. Chang, K.~Lee, and K.~Toutanova, ``Bert: Pre-training of deep
  bidirectional transformers for language understanding,'' \emph{arXiv preprint
  arXiv:1810.04805}, 2018.

\end{thebibliography}

\end{document}